\documentclass[10pt,twocolumn,letterpaper]{article}

\usepackage{subcaption}

\usepackage{cvpr}
\usepackage{times}

\usepackage{graphicx}
\graphicspath{{images/}}

\usepackage{amsmath}
\usepackage{amssymb}

\newcommand{\rulesep}{\unskip\ \textcolor{blue}{\vrule width 1.5pt\ }}
\def\etal{\emph{et al.}~}

\usepackage[pagebackref=true,breaklinks=true,colorlinks,bookmarks=false]{hyperref}

\cvprfinalcopy 

\def\cvprPaperID{601} 

\ifcvprfinal\fi
\begin{document}

\title{Hierarchical Boundary-Aware Neural Encoder for Video Captioning}

\author{Lorenzo Baraldi\quad Costantino Grana\quad Rita Cucchiara\\
University of Modena and Reggio Emilia\\
{\tt\small \{name.surname\}@unimore.it}
}

\maketitle
\thispagestyle{empty}

\begin{abstract}
The use of Recurrent Neural Networks for video captioning has recently gained a lot of attention, since they can be used both to encode the input video and to generate the corresponding description. In this paper, we present a recurrent video encoding scheme which can discover and leverage the hierarchical structure of the video. Unlike the classical encoder-decoder approach, in which a video is encoded continuously by a recurrent layer, we propose a novel LSTM cell which can identify discontinuity points between frames or segments and modify the temporal connections of the encoding layer accordingly. We evaluate our approach on three large-scale datasets: the Montreal Video Annotation dataset, the MPII Movie Description dataset and the Microsoft Video Description Corpus. Experiments show that our approach can discover appropriate hierarchical representations of input videos and improve the state of the art results on movie description datasets.
\end{abstract}

\section{Introduction}
Automatically describing a video in natural language is an important challenge for computer vision and machine learning. This task, called video captioning, is a crucial achievement towards machine intelligence and also the support of a number of potential applications. Indeed, bringing together vision and language, video captioning can be leveraged for video retrieval, to enhance content search on video sharing and streaming platforms, as well as to generate  automatic subtitles and to help visually impaired people to get an insight of the content of a video.

Before targeting videos, captioning has been tackled for images, where the task was that of generating a single sentence which described a static visual content~\cite{vinyals2015show,karpathy2015deep,xu2015show,vinyals2016show}. Later, image captioning approaches have been extended to short videos with a single action, object, or scene, initially using very similar approaches to image captioning, and then with solutions to account for the temporal evolution of the video~\cite{yao2015describing,sutskever2014sequence,yu2015video}. After having been applied to highly constrained or user generated videos~\cite{rohrbach2014coherent,chenacl11}, video captioning is moving to more complex and structured kinds of video, thanks to the spread of movie description datasets~\cite{torabi2015using, rohrbach15cvpr}.

\begin{figure}[tb]
\centering
\includegraphics[width=0.95\columnwidth]{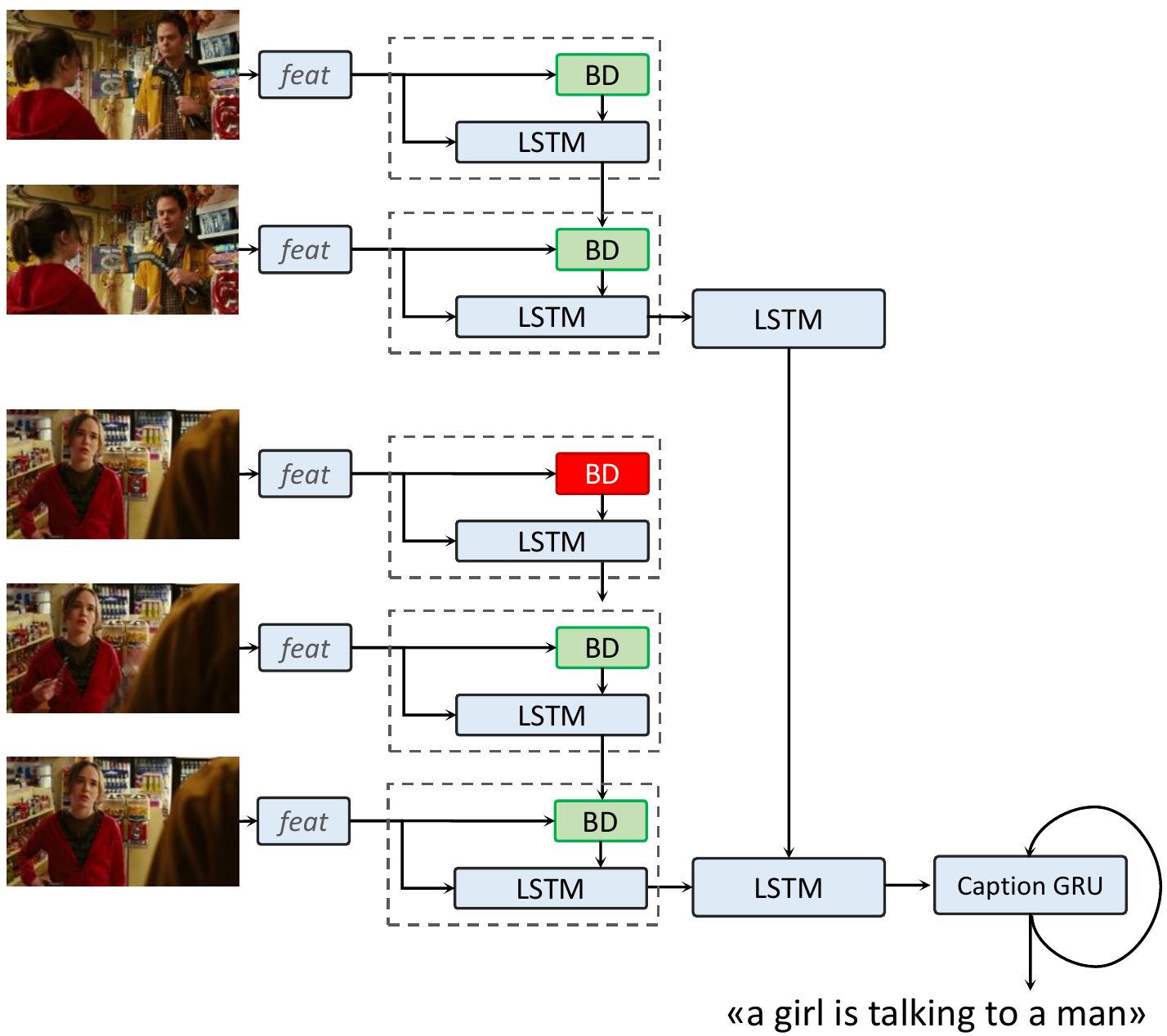}
\caption{We propose a novel video encoding network which can adaptively modify its structure to improve video captioning. Our Time Boundary-aware LSTM cell (depicted with dashed rectangles) extends the standard LSTM unit by adding a trainable boundary detector (BD), which can alter the temporal connections of the network depending on the input video.}
\label{fig:overview}
\end{figure}

So far, video captioning algorithms have relied on the use of Recurrent Neural Networks or Long Short-Term Memory (LSTM)~\cite{graves2013speech} layers, which can naturally deal with sequences of frames and, in principle, learn long-range temporal patterns. However, it has been proved that LSTMs show good learning capabilities on sequences which are between 30 and 80 frames long~\cite{yue2015beyond}, shorter than the ones used in video captioning. Furthermore, the plain nature of recurrent networks can not deal with the layered structure of videos. 

This is the case of edited video, such as movies. Long edited video can be segmented into short scenes, using Descriptive Video Services or with deep learning techniques~\cite{liu2013learning,baraldi2015deep}; however video scenes contain several shots that, although temporally consistent, have a different appearance. As an example, in Figure~\ref{fig:overview} two shots of a dialogue are depicted. In this case we want to prevent the network from mixing the memory of the two shots; conversely, if the network could be aware of the presence of a temporal boundary, it could reset its internal status creating a new output independent to the one of the previous shot. This also applies to user-generated video, where events can be composed by a sequence of actions in a single shot (e.g. a player runs and shoots the ball). An effective encoder should consider the temporal dependencies both intra-action and inter-actions.



In this paper, we propose a novel video encoding scheme for video captioning capable of identifying temporal discontinuities, like action or appearance changes, and exploiting them to get a better representation of the video. Figure~\ref{fig:overview} shows the hierarchical structure of our sequence-to-sequence architecture: frames, described by features computed by a CNN, enter into our time boundary-aware LSTM. The awareness of the presence of an appearance or action discontinuity automatically modifies the connectivity through time of the LSTM layer: the result is a variable length and adaptive encoding of the video, whose length and granularity depends on the input video itself. The outputs of the first boundary-aware layer are encoded through an additional recurrent layer into a fixed length vector, which is then used for generating the final caption through a Gated Recurrent Unit (GRU) layer.
The contributions of the paper are summarized below.
\begin{itemize}
	\item We present a new time boundary-aware LSTM cell: it can discover discontinuities in the input video and enables the encoding layer to modify its temporal connectivity, resetting its internal state and memory if required. The proposed cell incorporates a boundary detection module and encodes content and temporal structure in a trainable end-to-end layer.
	
	\item The time boundary-aware LSTM is used to build a hierarchical encoder for video captioning: to the best of our knowledge, this is the first proposal of a video captioning network which can learn to adapt its structure to input data.

	\item We test our approach on three large-scale movie description
	and video captioning datasets: 
	M-VAD~\cite{torabi2015using},
	MPII-MD~\cite{rohrbach15cvpr},
	and MSVD~\cite{chenacl11}.
	Our results significantly improve the state-of-the art on movie description, being competitive also on short user-generated video. We also investigate boundaries learned by our encoder, and show that it can discover appropriate decompositions of the input video. 
\end{itemize}

\section{Related Works}
\begin{figure*}[tb]
\centering
	\begin{minipage}{0.61\textwidth}
		\begin{subfigure}[b]{\textwidth}
			\includegraphics[width=\textwidth]{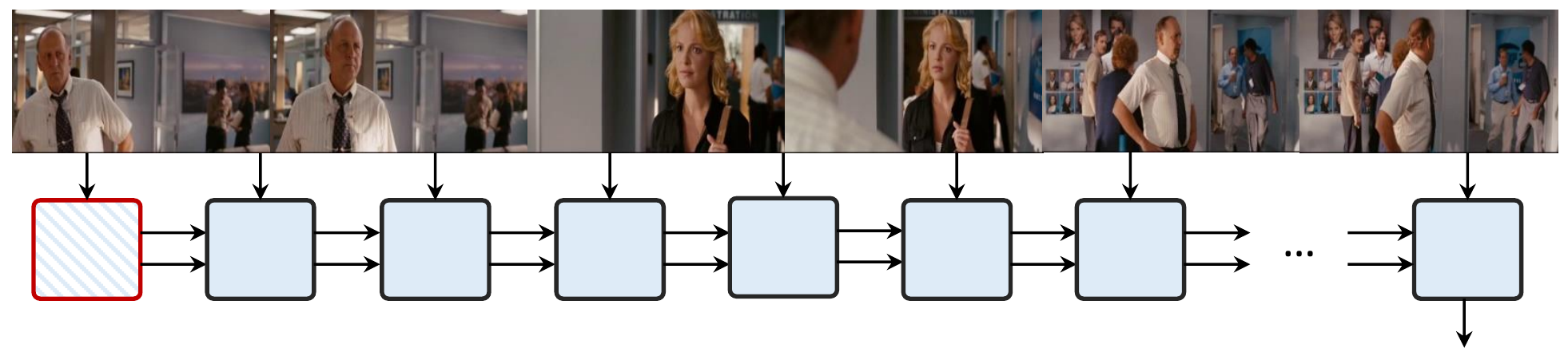}
			\caption{Traditional LSTM network}
			\label{fig:lstm}
		\end{subfigure}
		\vspace{0.05cm}
		
		\begin{subfigure}[b]{\textwidth}
			\includegraphics[width=\textwidth]{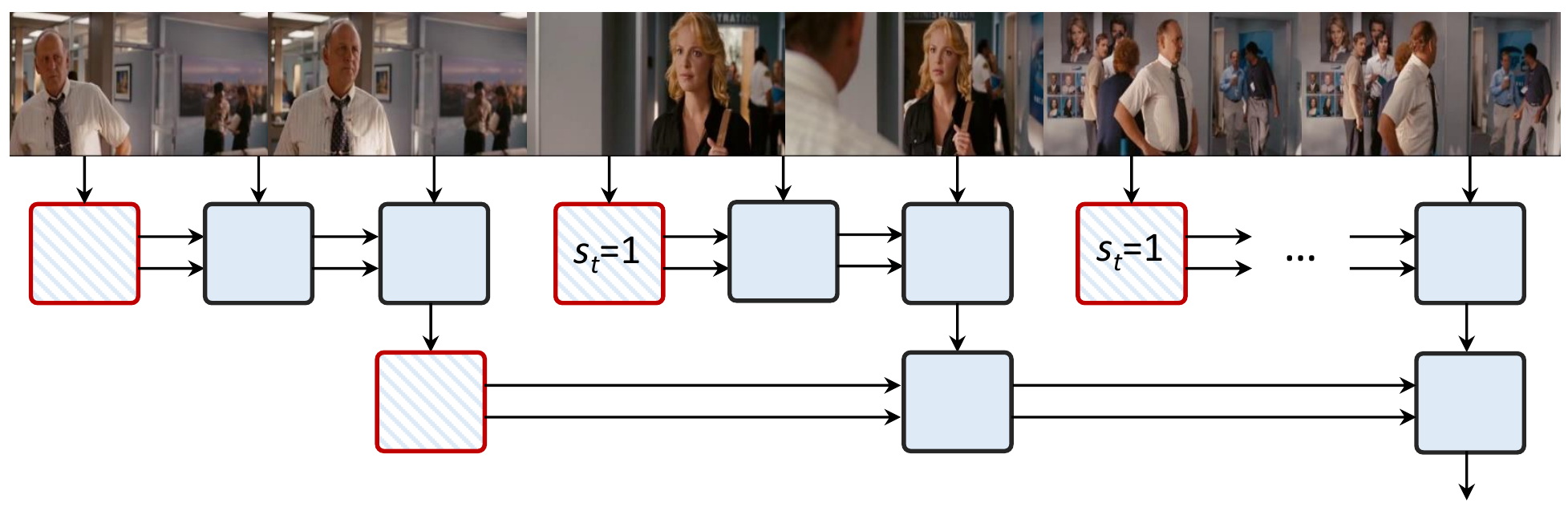}
			\caption{Time Boundary-aware LSTM network}
			\label{fig:hmlstm}
		\end{subfigure}
	\end{minipage}
	\quad \quad \quad
	\begin{minipage}{0.24\textwidth}
		\begin{subfigure}[b]{\textwidth}
			\includegraphics[width=\textwidth]{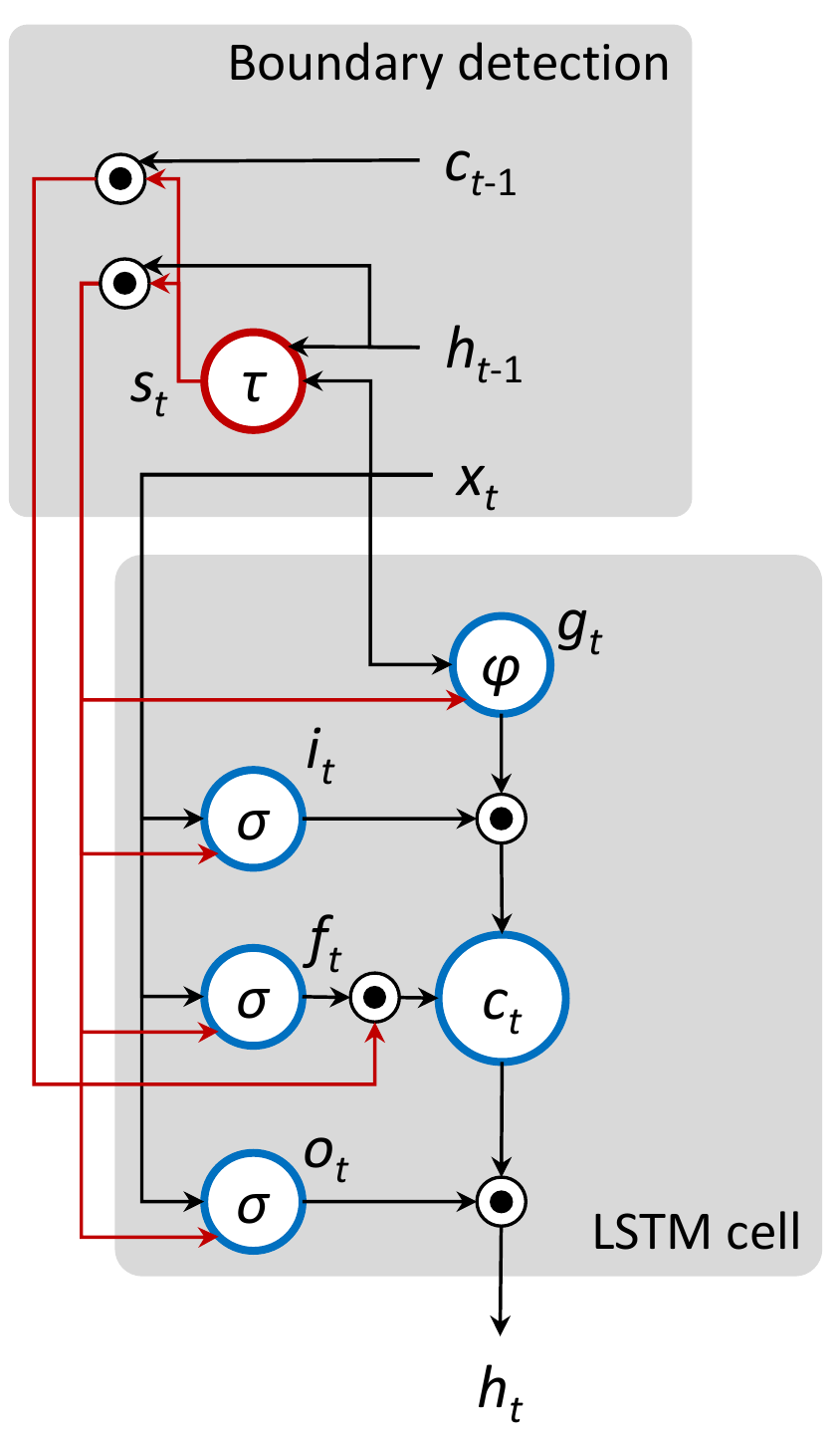}
			\caption{Time Boundary-aware cell}
			\label{fig:lstm_schema}
		\end{subfigure}		
	\end{minipage}
	
	\vspace{0.1cm}
	\caption{Comparison between a standard LSTM encoder and the Time Boundary-aware LSTM network, and schema of the Boundary-aware LSTM cell. The proposed video encoder can \textit{learn} to modify its temporal connections according to appearance or action changes which are found in the video: when a boundary is detected, the state of the LSTM is reinitialized and a representation of the ended segment is given to the output. Red dashed boxes represent LSTM units with reset state, black boxes stand for LSTM unit with modified states.}
	\label{fig:schemas}
\end{figure*}

Early captioning methods~\cite{guadarrama2013youtube2text,krishnamoorthy2013generating,thomason2014integrating} were based on the identification of (subject, verb, object) triplets with visual classifiers, and captions were generated through a language model which fitted predicted triplets to predefined sentence templates. Of course, template-based sentences can not satisfy the richness of natural language, and have limited ability to generalize to unseen data. For these reasons, research on image and video captioning has soon moved to the use of recurrent networks, which, given a vectored description of a visual content, could naturally deal with sequences of words~\cite{vinyals2015show,karpathy2015deep}.

In one of the first approaches to video captioning with recurrent networks, Venugopalan \etal\cite{venugopalan2014translating} used CNN features extracted from single frames, mean pooled them to represent the entire video, and then fed the resulting vector to a LSTM layer~\cite{graves2013speech} for the generation of the caption. The major drawback of this method is that it ignored the sequential nature of video, reducing the task of video captioning to a mere extension of image captioning. Therefore, many following works tried to develop more appropriate video encoding strategies. Donahue \etal\cite{donahue2015long}, for example, used a LSTM network to sequentially encode the input video, and then employed CRFs to get semantic tuples of activity, object, tool and location. A final LSTM layer translated the semantic tuple into a sentence.

Venugopalan \etal\cite{venugopalan2015sequence} proposed a completely neural architecture addressing both the video encoding stage and sentence decoding. They used a stacked LSTM to read the sequence of video frames, and a second LSTM, conditioned on the last hidden state of the first, to generate the corresponding caption. Interestingly, the LSTM parameters used in the two stages were shared. That was the first time the so-called \textit{sequence to sequence} approach, already applied to machine translation~\cite{sutskever2014sequence}, was used for video captioning. Other works have then followed this kind of approach, either by incorporating attentive mechanisms~\cite{yao2015describing} in the sentence decoder, by building a common visual-semantic emebedding~\cite{pan2015jointly}, or by adding external knowledge with language models~\cite{venugopalan16emnlp} or visual classifiers~\cite{rohrbach2015long}.

Recently, researches improved both the components of the encoder-decoder approach by significantly changing their structure. Yu \etal\cite{yu2015video} focused on the sentence decoder, and proposed a hierarchical model containing a sentence and a paragraph generator: short sentences are produced by a Gated Recurrent Unit (GRU) layer~\cite{cho2014learning} conditioned on video features, while another recurrent layer is in charge of generating paragraphs by combining sentence vectors and contextual information. The paragraph generator can therefore captures inter-sentence dependencies and generate a sequence of related and consecutive sentences. In this paper, as in their proposal, we adopt a final GRU layer for the generation of the caption.

In contrast, Pan \etal\cite{pan2015hierarchical} targeted the video encoding stage, by proposing a hierarchical recurrent video encoder. Their proposal tries to abstract visual features at different time scales and granularities, by processing frames of the video in a way similar to a convolutional operation applied in the time dimension. A LSTM is applied to small overlapped video chunks, in a sliding window fashion: this results in a sequence of vectors, which are then forwarded to a second recurrent layer, or processed by the decoder LSTM through a soft attention mechanism~\cite{bahdanau2014neural}. 

Also in this paper, we focus on the video encoding stage. However, instead of building an hand-crafted variation of the plain LSTM layer as in~\cite{pan2015hierarchical}, we propose a recurrent network which can learn to adapt its temporal structure to input data. Our strategy, contrary to the sliding window approach, also ensures that the cell memory encoding each chunk always contains homogeneous information. The idea of leveraging segment-level features has been investigated in natural language processing~\cite{chung2016hierarchical}, action recognition~\cite{tang2012learning,song2013action,pirsiavash2014parsing,lan2015action} and event detection~\cite{wang2016temporal}. Our network is the first proposal which exploits temporal segments in video captioning.

\section{Method}
Given an input video, we propose a recurrent video encoder which takes as input a sequence of visual features $(\mathbf{x}_1, \mathbf{x}_2, ..., \mathbf{x}_n)$ and outputs a sequence of vectors $(\mathbf{s}_1, \mathbf{s}_2, ..., \mathbf{s}_m)$ as the representation for the whole video. 
In our encoder, the connectivity schema of the layer varies with respect to both the current input and the hidden state, so it is thought as an \textit{activation} instead of being a non learnable hyperparameter. 

To this aim, we define a time boundary-aware recurrent cell, which can modify the layer connectivity through time: when an appearance or action change is estimated, the hidden state and the cell memory are reinitialized, and at the end of a segment the hidden state of the layer is given to the output, as a summary of the detected segment. This ensures that the input data following a time boundary are not influenced by those seen before the boundary, and generates a hierarchical representation of the video in which each chunk is composed by homogeneous frames. Figures~\ref{fig:lstm} and~\ref{fig:hmlstm} show the temporal connections determined by the boundary detector in a sample case, compared to those of a plain LSTM encoder.

The proposed time boundary-aware recurrent cell is built on top of a Long Short-Term Memory (LSTM) unit, which has been shown to be particularly suited to video encoding, since it is known to learn patterns with wide temporal dependencies. At its core there is a memory cell $\mathbf{c}_t$ which maintains the history of the inputs observed up to a timestep. Update operations on the memory cell are modulated by three gates $\mathbf{i}_t$, $\mathbf{f}_t$ and $\mathbf{o}_t$, which are all computed as a combination of the current input $\mathbf{x}_t$ and of the previous hidden state $\mathbf{h}_{t-1}$, followed by a sigmoid activation. The input gate $\mathbf{i}_t$ controls how the current input should be added to the memory cell; the forget gate $\mathbf{f}_t$ is used to control what the cell will forget from the previous memory $\mathbf{c}_{t-1}$, and the output gate $\mathbf{o}_t$ controls whether the current memory cell should be passed as output. 

At each timestep, we select whether to transfer the hidden state and memory cell content to the next timestep or to reinitialize them, interrupting the seamless update and processing of the input sequence. This depends on a time boundary detection unit, which allows our encoder to independently process variable length chunks of the input video. The boundaries of each chunk are given by a learnable function which depends on the input, and are not set in advance.

Formally, the boundary detector $s_t \in \{ 0, 1 \}$ is computed as a linear combination of the current input and of the hidden state, followed by a function $\tau$ which is the composition of a sigmoid and a step function:
\begin{align}
s_t &= \tau ( \mathbf{v}_s^T \cdot ( W_{si} \mathbf{x}_t + W_{sh} \mathbf{h}_{t-1} + \mathbf{b}_s )) \\
\tau(x) &= \begin{cases} 1, & \mbox{if } \sigma(x) > 0.5 \\ 0, & \mbox{otherwise} \end{cases}
\label{eq:deterministic_step}
\end{align}
where $\mathbf{v}_s^T$ is a learnable row vector and $W_{sh}, \mathbf{b}_s$ are learned weights and biases.

Given the current boundary detection $s_t$, before applying the memory unit update equations, the following substitutions are applied to transfer or reinitialize the network hidden state and memory cell at the beginning of a new segment, according to $s_t$:
\begin{align}
\mathbf{h}_{t-1} \leftarrow \mathbf{h}_{t-1} \cdot (1-s_t) \\
\mathbf{c}_{t-1} \leftarrow \mathbf{c}_{t-1} \cdot (1-s_t).
\end{align}
The resulting state and memory are now employed to recompute the gates values, which will in turn be used for advancing to the next time step.
The encoder produces an output only at the end of a segment. If $s_{t}=1$, indeed, the hidden state of timestep $t-1$ is passed to the next layer. 

Many LSTM architectures have been proposed~\cite{hochreiter1997long,gers2000learning,schmidhuber2007training,graves2013speech}, and all are slightly different in their structure and activation functions, even though they all share the presence of additive memory cells and gates. In our case, we apply the following equations~\cite{graves2013speech}
\begin{align}
\mathbf{i}_t &= \sigma(W_{ix} \mathbf{x}_t + W_{ih} \mathbf{h}_{t-1} + \mathbf{b}_i) \\
\mathbf{f}_t &= \sigma(W_{fx} \mathbf{x}_t + W_{fh} \mathbf{h}_{t-1} + \mathbf{b}_f) \\
\mathbf{g}_t &= \phi(W_{gx} \mathbf{x}_t + W_{gh} \mathbf{h}_{t-1} + \mathbf{b}_g) \\
\mathbf{c}_t &= \mathbf{f}_t \odot \mathbf{c}_{t-1} + \mathbf{i}_t \odot \mathbf{g}_t \\
\mathbf{o}_t &= \phi(W_{fx} \mathbf{x}_t + W_{fh} \mathbf{h}_{t-1} + \mathbf{b}_f) \\
\mathbf{h}_t &= \mathbf{o}_t \odot \phi(\mathbf{c}_t)
\label{eq:lstm}
\end{align}
where $\odot$ denotes the element-wise Hadamard product, $\sigma$ is the sigmoid function, $\phi$ is the hyperbolic tangent \texttt{tanh}, $W_{*}$ are learned weight matrices and $\mathbf{b}_*$ are learned biases vectors. The internal state $\mathbf{h}$ and memory cell $\mathbf{c}$ are initialized to zero. Figure~\ref{fig:lstm_schema} shows a schema of the proposed time boundary-aware cell.



A recurrent layer which follows the equations reported above will produce a variable length set of outputs $(\mathbf{s}_1, \mathbf{s}_2, ..., \mathbf{s}_m)$, where $m$ is the number of detected segments. Each of these outputs conceptually summarizes the content of a detected segment inside the video. This set of outputs is passed to another recurrent layer, thus building a hierarchical representation of the video. To this end, we fed the output of the boundary-aware encoder to an additional LSTM layer, whose last hidden state can be used as the feature vector for the entire video.

Existing approaches to video encoding add more non-linearity to LSTM architectures by stacking together more layers~\cite{venugopalan2015sequence}, or by building a hierarchical architecture in which a lower level encodes fixed length chunks, while a higher level is in charge of composing these encoded chunks to get the final video representation~\cite{pan2015hierarchical}. Our proposal, while keeping a completely neural architecture, enables the encoder to both produce variable length chunks, based on the input data characteristics, and to encode them in a hierarchical structure.


\subsection{Training}
Due to the presence of a binary variable which influences the temporal structure of the video encoder, special training expedients are required.

First of all, the boundary detector $s_t$ is treated at training time as a stochastic neuron~\cite{raiko2014techniques}. In particular, we introduce a stochastic version of function $\tau(x)$ (Eq.~\ref{eq:deterministic_step}), in which its ouput is sampled from a uniform distribution conditioned on $\sigma(x)$. Formally, during the forward pass of the training phase, $\tau$ is computed as
\begin{equation}
\tau(x) = \mathbf{1}_{\sigma(x) > z} \mbox{, with } z\sim U\left[0,1\right], \mbox{forward pass}
\end{equation}
where $U\left[0,1\right]$ is a uniform distribution over the interval $\left[0,1\right]$ and $\mathbf{1}$ is the indicator function. This ensures that $s_t$ is stochastic, and its probability of being 0 or 1 is proportional to the value of a sigmoid applied to the input of $\tau$. 

In the backward pass, since the derivative of the step function is zero almost anywhere, the standard back propagation would no longer be applicable. To solve this issue, we employ an estimator of the step function as suggested by Bengio \etal\cite{bengio2013estimating}. The idea is that discrete operations can be used in the forward pass if a differentiable approximation is used in the backward one. In our case, we approximate the step function with the identity function, which has shown good performances~\cite{bengio2013estimating}. Being $\tau$ the composition of a sigmoid and a step function, the derivative of $\tau$ used in backward is simply the derivative of the sigmoid function.
\begin{equation}
\frac{\partial \tau}{\partial x}(x) = \sigma(x)(1-\sigma(x)), \mbox{backward pass}
\end{equation}
At test time, the deterministic version of the step function (Eq.~\ref{eq:deterministic_step}) is used. In this way the number of detected segments is stochastic during training and deterministic during test.

\subsection{Sentence generation}
\label{sec:sentence_generation}
Once the representation of the video has been computed, the description of the video is generated through a decoder network, following the encoder-decoder scheme~\cite{yao2015describing,venugopalan2015sequence,pan2015hierarchical}.

Given a video vector $\mathbf{v}$ and a sentence $(\mathbf{y}_0, \mathbf{y}_1, ..., \mathbf{y}_T)$, encoded with one-hot vectors (1-of-$N$ encoding, where $N$ is the size of the vocabulary),  our decoder is conditioned step by step on the first $t$ words of the caption and on the corresponding video descriptor, and is trained to produce the next word of the caption. The objective function which we optimize is the log-likelihood of correct words over the sequence
\begin{equation}
\max_{\mathbf{w}} \sum_{t=1}^T \log \text{Pr} (\mathbf{y}_t | \mathbf{y}_{t-1}, \mathbf{y}_{t-2}, ..., \mathbf{y}_0, \mathbf{v})
\end{equation}
where $\mathbf{w}$ are all the parameters of the encoder-decoder model. The probability of a word is modeled via a softmax layer applied on the output of the decoder. To reduce the dimensionality of the decoder, a linear embedding transformation is used to project one-hot word vectors into the input space of the decoder and, viceversa, to project the output of the decoder to the dictionary space.
\begin{equation}
\text{Pr} (\mathbf{y}_t | \mathbf{y}_{t-1}, \mathbf{y}_{t-2}, ..., \mathbf{y}_0, \mathbf{v}) \propto \exp (\mathbf{y}_t^T W_p \mathbf{p}_t)
\end{equation}
$W_p$ is a matrix for transforming the decoder output space to the word space and $\mathbf{p}_t$ is the output of the decoder, computed with a Gated Recurrent Unit (GRU)~\cite{cho2014learning} layer.

The output of the GRU layer, at each timestep, is modeled via two sigmoid gates: a reset gate ($\mathbf{r}_t$), which determines whether the previous hidden state should be dropped to generate the next outputs, and an update gate ($\mathbf{z}_t$) which controls how much information of the previous hidden state should be preserved:
\begin{align}
\mathbf{z}_t &= \sigma(W_{zy} W_{w} \mathbf{y}_t + W_{zv} \mathbf{v} + W_{zh} \mathbf{p}_{t-1} + \mathbf{b}_i) \\
\mathbf{r}_t &= \sigma(W_{ry} W_{w} \mathbf{y}_t + W_{rv} \mathbf{v} + W_{rh} \mathbf{p}_{t-1} + \mathbf{b}_f).
\end{align}
Exploiting the values of the above gates, the output of the decoder GRU is computed as:
\begin{align}
\tilde{\mathbf{h}}_t &= \phi(W_{hy} W_{w} \mathbf{y}_t + W_{hv} \mathbf{v} + W_{hh} (\mathbf{r}_t \odot \mathbf{p}_{t-1}) + \mathbf{b}_f)\\
\mathbf{p}_t &= (1-\mathbf{z}_t) \odot \tilde{\mathbf{h}}_{t-1} + \mathbf{z}_t \odot \mathbf{p}_t
\end{align}
where $W_*$ and $\mathbf{b}_*$ are learned weights and biases and $W_w$ transforms the one-hot encoding of words to a dense lower dimensional embedding. Again, $\odot$ denotes the element-wise product, $\sigma$ is the sigmoid function and $\phi$ is the hyperbolic tangent.


\section{Experimental setup}
Evaluation is carried out on three large-scale datasets for video captioning, one containing user-generated videos, and the other two specifically built for movie description.

\subsection{Datasets}
\textbf{Montreal Video Annotation dataset (M-VAD)}~The Montreal Video Annotation dataset~\cite{torabi2015using} is a large-scale video description dataset based on Descriptive Video Service (DVS). DVS, or Audio Descriptions, are audio tracks describing the visual elements of a movie, produced to help visually impaired people. The dataset consists of 84.6 hours of video from 92 Hollywood movies, for a total of 46,523 video clips, each automatically aligned with with a single description. We use the standard splits provided in~\cite{torabi2015using}, which consists of 36,921 training samples, 4,651 validation samples and 4,951 test samples. 

\textbf{MPII Movie Description dataset (MPII-MD)}~The MPII Movie Description dataset~\cite{rohrbach15cvpr} has been built in a way similar to M-VAD, even though in this case the alignment between video snippets and descriptions is more correct, since it has been manually corrected. The dataset contains a parallel corpus of over 68K sentences and video snippets from 94
HD movies, obtained from scripts and Audio Descriptions. Following the splits provided by the authors, the dataset contains 56,861 train samples, 4,930 validation samples and 6,584 test samples.

\textbf{Microsoft Video Description Corpus (MSVD)}~The Microsoft Video Description Corpus~\cite{chenacl11} contains 2,089 Youtube video clips, labeled with 85K English descriptions collected by Amazon Mechanical Turkers. The dataset was initially conceived to contain multi-lingual descriptions; however, we only consider captions in the English language. As done in previous works~\cite{guadarrama2013youtube2text, venugopalan2014translating}, we split the dataset in contiguous groups of videos by index number: 1,200 for training, 100 for validation and 670 for test. This dataset mainly contains short video clips with a single action, an is therefore less appropriate than M-VAD and MPII-MD to evaluate the effectiveness of our method in identifying the video structure.

\subsection{Metrics}
We employ four popular metrics for evaluation: BLEU~\cite{papineni2002bleu}, $\text{ROUGE}_L$~\cite{lin2004rouge}, METEOR~\cite{banerjee2005meteor} and CIDEr~\cite{vedantam2015cider}. BLEU is a form of precision of word n-grams
between predicted and ground-truth sentences. As done in previous works, we evaluate our predictions with BLEU using four-grams. $\text{ROUGE}_L$ computes an F-measure with a recall bias using a longest common subsequence technique. METEOR, instead, scores captions by aligning them to one or more ground truths. Alignments are based on exact, stem, synonym, and paraphrase matches between words and phrases, therefore METEOR is more semantically adequate than BLEU and $\text{ROUGE}_L$. CIDEr, finally, computes the average cosine similarity between n-grams found in the generated caption and those found in reference sentences, weighting them using TF-IDF. The authors of CIDEr~\cite{vedantam2015cider} reported that CIDEr and METEOR are always more accurate, especially when the number of reference captions is low.

To ensure a fair evaluation, we use the Microsoft CoCo evaluation toolkit\footnote{\url{https://github.com/tylin/coco-caption}} to compute all scores, as done in previous video captioning works~\cite{yu2015video,pan2015hierarchical}.

\subsection{Preprocessing and training details}
We extract static appearance as well as motion features from input videos of all datasets. To encode video appearance, we use the ResNet50 model~\cite{He2015} trained on the Imagenet dataset~\cite{russakovsky2015imagenet}, and compute a descriptor every 5 frames. For motion, we employ the C3D network~\cite{tran2015learning} (trained on the Sports-1M dataset~\cite{KarpathyCVPR14}): this model outputs a fixed length feature vector every 16 frames, which encodes motion features computed around the middle frame of the window. To maintain the same granularity used for appearance, we sample 16 frames long, partially overlapped, windows with a stride of 5 frames. In both cases, we use the activations from the penultimate layer of the network, which leads to a 2,048+4,096-dimensional feature vector. Instead of directly inputting visual features into our model, we learn a linear embedding as the input of the model.

Ground truth descriptions are converted to lower case and tokenized after having removed punctuation characters. We retain only words which appear at least five times in a dataset. This yields a vocabulary of 6,090 words for the M-VAD dataset, 7,198 words for MPII-MD and 4,215 words for MSVD. During training, we add a begin-of-sentence $<$BOS$>$ tag at the beginning of the caption, and end-of-sentence tag $<$EOS$>$ at its end, so that our model can deal with captions with variable length. At test time, the decoder RNN is given a $<$BOS$>$ tag as input for the first timestep, then the most probable word according to the predicted distribution is sampled and given as input for the next timestep, until a $<$EOS$>$ tag is predicted.

Training is performed by minimizing the log-likelihood loss with the Adadelta optimizer, with a learning rate of 1.0 and decay parameters $\rho = 0.95$ and $\epsilon = 10 \times 10^{-7}$, which generally show good performance. We set the mini-batch size to 128. To regularize the training and avoid overfitting, we apply the well known regularization technique Dropout~\cite{srivastava2014dropout} with retain probability 0.5 on the input and output of the encoding LSTMs, as suggested by Zaremba \etal\cite{zaremba2014recurrent}.

Embeddings for video features and words have all size 512, while the size of all recurrent hidden state is empirically set to 1024. Regarding initializiation, we used the gaussian initialization suggested by Glorot \etal\cite{glorot2010understanding} for weight matrices applied to inputs, and orthogonal initialization for weight matrices applied to internal states. Embedding matrices were also initialized according to~\cite{glorot2010understanding}, and all biases were initialized to zero.

We train the model for 100 epochs, or until the loss improvement over the validation set stops.
The source code of model has been written using Theano, and is made publicly available\footnote{\url{http://imagelab.ing.unimore.it/video_captioning}}.

\section{Results and Discussion}
\begin{table}[tb]
\begin{center}
\small
\begin{tabular}{l|c}
\hline
Model & METEOR \\
\hline
SA-GoogleNet+3D-CNN~\cite{yao2015describing} 			& 4.1 \\
HRNE~\cite{pan2015hierarchical} 									& 5.8 \\
S2VT-RGB(VGG)~\cite{venugopalan2015sequence} 			& 6.7 \\
HRNE with attention~\cite{pan2015hierarchical} 		& 6.8 \\
Venugopalan \etal\cite{venugopalan16emnlp} 				& 6.8 \\
\hline
LSTM encoder (C3D+ResNet)													& 6.7 \\
Double-layer LSTM encoder (C3D+ResNet)						& 6.7 \\
Boundary encoder on shots 												& 7.1 \\
Boundary-aware encoder (C3D+ResNet)								& \textbf{7.3}  \\
\hline
\end{tabular}
\end{center}
\caption{Experiment results on the M-VAD dataset.}
\label{tab:mvad}
\end{table}

\begin{table}[tb]
\begin{center}
\resizebox{\columnwidth}{!}{
\begin{tabular}{l|c|c|c|c}
\hline
Model & CIDEr & B@4 & $\text{R}_L$ & M \\
\hline
SMT (best variant)~\cite{rohrbach15cvpr} 					& 8.1 	& 0.5 	& 13.2 	& 5.6 \\
SA-GoogleNet+3D-CNN~\cite{yao2015describing} 			& - 		& - 		& - 		& 5.7 \\
Venugopalan \etal\cite{venugopalan16emnlp} 				& - 		& - 		& - 		& 6.8 \\
Rohrbach \textit{et al.}~\cite{rohrbach2015long} 	& 10.0 	& \textbf{0.8} 	& 16.0 & \textbf{7.0} \\
\hline
LSTM encoder (C3D+ResNet)								& 10.5 & 0.7 & 16.1 & 6.4 \\
Double-layer LSTM encoder (C3D+ResNet)	&	10.6 & 0.6 & 16.5	& 6.7 \\
Boundary encoder on shots & 10.3 & 0.7 & 16.3 & 6.6 \\
Boundary-aware encoder (C3D+ResNet) & \textbf{10.8} & \textbf{0.8} & \textbf{16.7} & \textbf{7.0}  \\
\hline
\end{tabular}
}
\end{center}
\caption{Experiment results on the MPII-MD dataset.}
\label{tab:mpiimd}
\end{table}

\begin{figure*}[tb]
\centering
\includegraphics[width=0.11\linewidth]{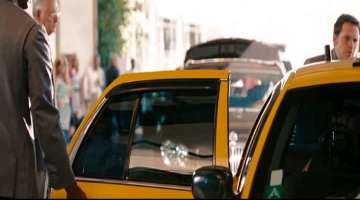}
\includegraphics[width=0.11\linewidth]{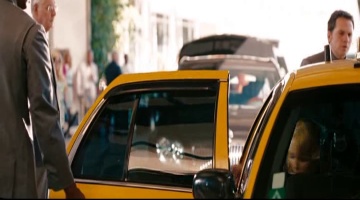}
\includegraphics[width=0.11\linewidth]{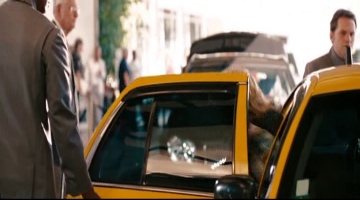}
\includegraphics[width=0.11\linewidth]{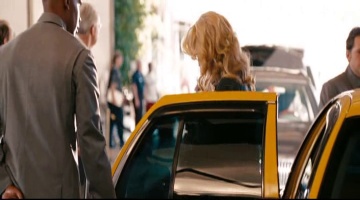}
\includegraphics[width=0.11\linewidth]{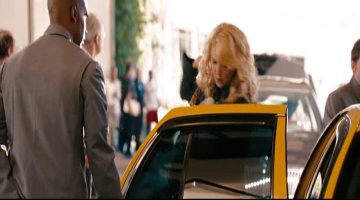}
\includegraphics[width=0.11\linewidth]{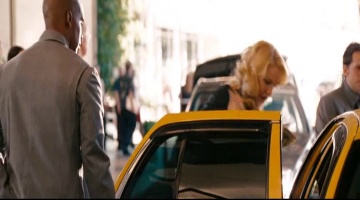}
\rulesep
\includegraphics[width=0.11\linewidth]{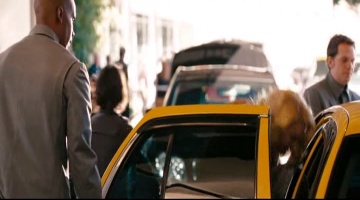}
\includegraphics[width=0.11\linewidth]{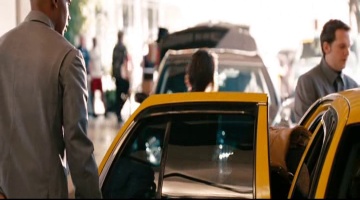} \\
\small{
\textbf{GT:} She gets out. \\
\textbf{LSTM encoder:} Someone stops.\\
\textbf{BA encoder (ours):} Someone gets out of the car.
}
\vspace{0.25cm}

\includegraphics[width=0.11\linewidth]{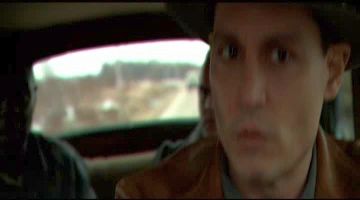}
\includegraphics[width=0.11\linewidth]{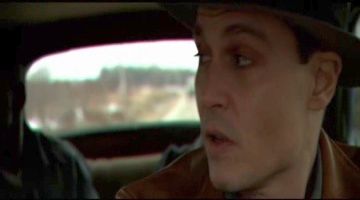}
\rulesep
\includegraphics[width=0.11\linewidth]{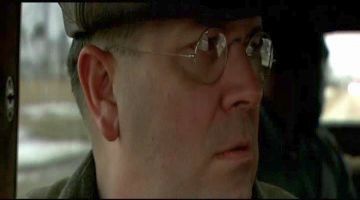}
\includegraphics[width=0.11\linewidth]{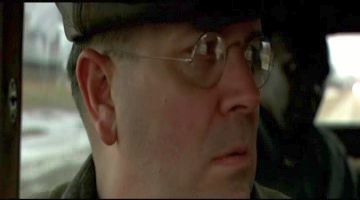}
\rulesep
\includegraphics[width=0.11\linewidth]{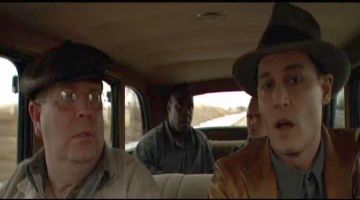}
\includegraphics[width=0.11\linewidth]{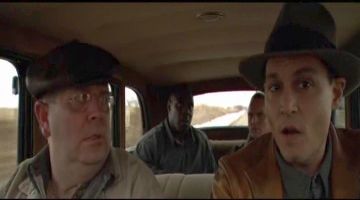}
\includegraphics[width=0.11\linewidth]{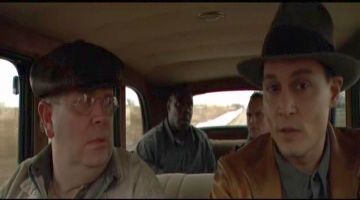}
\rulesep
\includegraphics[width=0.11\linewidth]{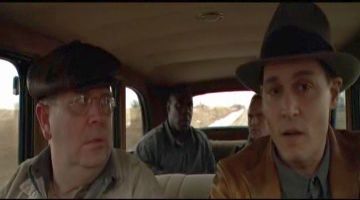} \\
\small{
\textbf{GT:} Shakes his head. \\
\textbf{LSTM encoder:} Someone gives her gaze. \\
\textbf{BA encoder (ours):} Someone looks at someone who shakes his head.
}
\vspace{0.25cm}

\includegraphics[width=0.11\linewidth]{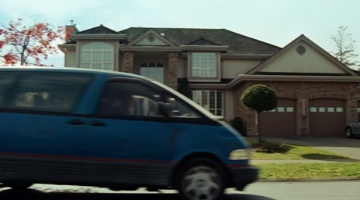}
\includegraphics[width=0.11\linewidth]{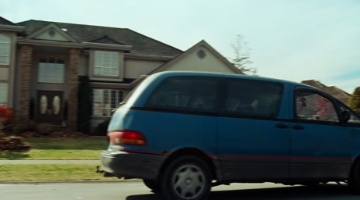}
\includegraphics[width=0.11\linewidth]{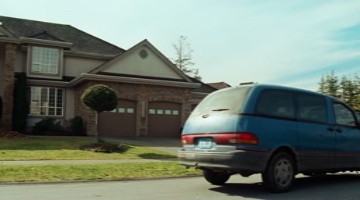}
\includegraphics[width=0.11\linewidth]{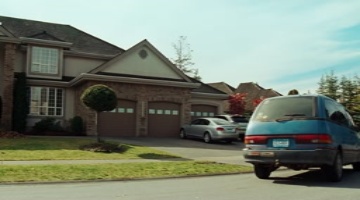}
\includegraphics[width=0.11\linewidth]{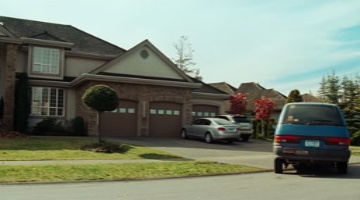}
\includegraphics[width=0.11\linewidth]{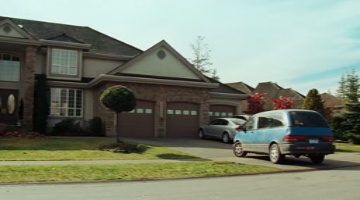}
\includegraphics[width=0.11\linewidth]{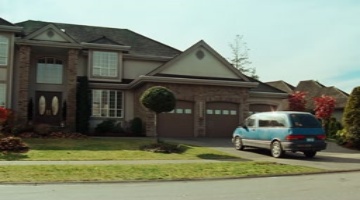}
\rulesep
\includegraphics[width=0.11\linewidth]{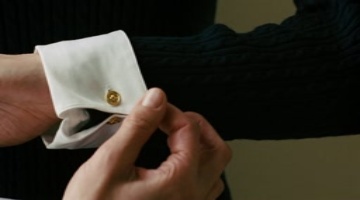} \\
\small{
\textbf{GT:} He slows down in front of one house with a garage and box tree on the front. \\
\textbf{LSTM encoder:} Someone gets out of the car and walks out of the house.\\
\textbf{BA encoder (ours):} Someone drives up to the house.
}

\caption{Example results on the M-VAD and MPII-MD dataset. Blue vertical lines represent an activation of the boundary detector in the LSTM cell.}
\label{fig:samples}
\end{figure*}

\subsection{Comparison with the state of the art}
On the M-VAD dataset we compare our method with four recent proposals: Temporal attention (SA)~\cite{yao2015describing}, S2VT~\cite{venugopalan2015sequence}, HRNE~\cite{pan2015hierarchical}, and the approach from Venugopalan \etal\cite{venugopalan16emnlp}. SA employed a LSTM decoder with a temporal attention mechanism over features extracted from GoogleNet~\cite{szegedy2015going} and from a 3D spatio-temporal CNN. S2VT, instead, used stacked LSTMs both for the encoder and the decoder stage, and frame-level features extracted from the VGG model. HRNE runs a LSTM on short video chunks, in a sliding window fashion, and the decoder selectively attends to the resulting set of vectors, optionally through a soft attention mechanism; the approach from Venugopalan \etal\cite{venugopalan16emnlp}, finally, focuses on the language model by adding knowledge from text corpora to the S2VT architecture.

Table~\ref{tab:mvad} shows the results on this dataset. As done in most of the previous video captioning works, we use METEOR as the main comparison metric. Firstly, to investigate the role of the boundary-aware encoder, we compare its performance against that of a single LSTM layer and that of a 2-layers LSTM encoder, trained using the same features and same hyperparameters. In this case, the last hidden state is used as the video vector for the GRU decoder. These baselines achieve a 6.7\% METEOR, while using the proposed encoder significantly increases performance, yielding to a 7.3\% METEOR which corresponds to an improvement of $0.6\%$.
This result also outperforms the most recent state-of-the-art method by a margin of $\frac{7.3 - 6.8}{6.8}= 7.35\%$. For reference, our method achieves a 0.9\% BLEU-4, 17.1\% $\text{ROUGE}_L$ and 10.4\% CIDER.

On the MPII-MD dataset, we again consider Temporal attention (SA)~\cite{yao2015describing}, S2VT~\cite{venugopalan2015sequence}, as well as the approach from Venugopalan \etal\cite{venugopalan16emnlp}. We also include two other references, which are applicable to this dataset: the Statistical Machine Translation (SMT) approach in~\cite{rohrbach15cvpr} and the work by Rohrbach \etal\cite{rohrbach2015long}, which exploits visual classifiers trained on visual labels extracted from captions.

The performance of these approaches and that of our solution is reported in Table~\ref{tab:mpiimd}. We observe that our approach is able to exceed the current state of the art on the CIDEr and $\text{ROUGE}_L$ metrics, while we achieve almost the same performance of the semantic approach of~\cite{rohrbach2015long} according to BLEU-4 and METEOR, without exploiting the semantics of captions and building concept classifiers. For reference, \cite{venugopalan2015sequence} reported a $7.1\%$ METEOR on this dataset. As for the M-VAD dataset, we also compare our solution to the baseline with a single LSTM layer: in this case, the improvement of the boundary-aware encoder is $0.6\%$ METEOR.

In Figure~\ref{fig:samples} we present a few examples of descriptions generated by our model on clips from the M-VAD and MPII-MD. We notice that the results obtained with the Boundary-aware encoder are generally better than those of the plain LSTM encoder, which is consistent with the results reported in Table~\ref{tab:mvad} and \ref{tab:mpiimd}.

As an additional test, we apply our method on MSVD, a common dataset for video captioning in which the hierarchical video structure is absent. The purpose, in this case, is to investigate whether our strategy impacts negatively when there is no structure in the video.

We compare our approach on MSVD with five state of the art approaches for video captioning: Temporal attention (SA)~\cite{yao2015describing}, LSTM-YT~\cite{venugopalan2014translating}, S2VT~\cite{venugopalan2015sequence}, LSTM-E~\cite{pan2015jointly} and HRNE~\cite{pan2015hierarchical}. LSTM-YT used a mean pool strategy on frame-level CNN features to encode the input video, while the caption was generated by a LSTM layer. LSTM-E, instead, proposed a visual-semantic embedding in which video descriptors and captions were projected, by maximizing distances between the projection of a video and that of its corresponding captions. As it can be noticed in Table~\ref{tab:msvd}, our method improves over plain techniques and can achieve competitive results. It is also worth noting that the attentive mechanism used in~\cite{pan2015hierarchical} could be integrated in our method, and potentially improve performance.

\begin{figure}[tb]
\begin{center}

\begin{minipage}{0.2\linewidth}
\includegraphics[width=0.9\linewidth]{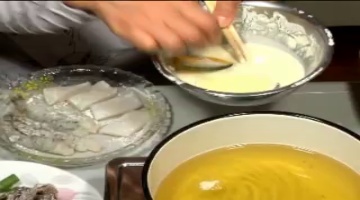}
\includegraphics[width=0.9\linewidth]{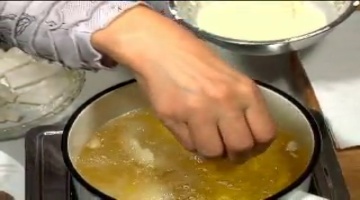}
\end{minipage}
\begin{minipage}{0.75\linewidth}
\small{
\textbf{GT:} A woman dips a shrimp in batter.\\
\textbf{HRNE~\cite{pan2015hierarchical}:} A woman is cooking. \\
\textbf{BA encoder (ours):} A woman is adding ingredients to a bowl of food.
}
\end{minipage}
\vspace{0.25cm}

\begin{minipage}{0.2\linewidth}
\includegraphics[width=0.9\linewidth]{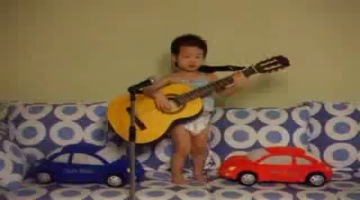}
\includegraphics[width=0.9\linewidth]{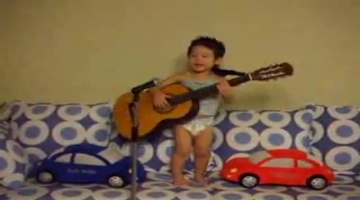}
\end{minipage}
\begin{minipage}{0.75\linewidth}
\small{
\textbf{GT:} A boy is playing a guitar.\\
\textbf{HRNE~\cite{pan2015hierarchical}:} A man is playing a guitar. \\
\textbf{BA encoder (ours):} A boy is playing guitar.
}
\end{minipage}
\vspace{0.25cm}

\begin{minipage}{0.2\linewidth}
\includegraphics[width=0.9\linewidth]{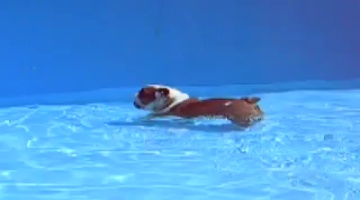}
\includegraphics[width=0.9\linewidth]{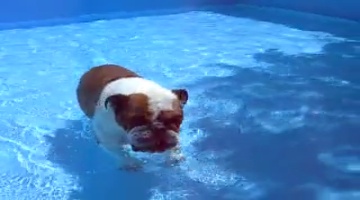}
\end{minipage}
\begin{minipage}{0.75\linewidth}
\small{
\textbf{GT:} A dog is swimming in a pool.\\
\textbf{HRNE~\cite{pan2015hierarchical}:} A dog is swimming. \\
\textbf{BA encoder (ours):} A dog is swimming in the pool.
}
\end{minipage}
\vspace{0.25cm}

\end{center}
\caption{Example results on the MSVD dataset.}
\label{fig:samples_msvd}
\end{figure}

Figure~\ref{fig:samples_msvd} reports some sample results on MSVD, comparing captions generated by our approach to those from the state of the art approach in~\cite{pan2015hierarchical}. As it can be seen, even though our method has not been conceived for videos lacking structure, it is still capable of generating accurate captions even in some difficult cases.

\begin{table}[tb]
\begin{center}
\small
			\begin{tabular}{l|c|c|c}
				\hline
				Model 																				& B@4 	& M 		& C 		\\
				\hline
				SA-GoogleNet+3D-CNN~\cite{yao2015describing}	& 41.9	& 29.6	& -			\\
				LSTM-YT~\cite{venugopalan2014translating} 		& 33.3	& 29.1 	& - 		\\
				S2VT~\cite{venugopalan2015sequence}						& -			& 29.8	& - 		\\
				LSTM-E~\cite{pan2015jointly}									& 45.3	& 31.0	& - 		\\
				HRNE~\cite{pan2015hierarchical}								& 46.7	& 33.9	& -			\\
				\hline
				Boundary-aware encoder									  		& 42.5	& 32.4 		& 63.5 		\\
				\hline
			\end{tabular}
\end{center}
	\caption{Experiment results on the MSVD dataset.}
	\label{tab:msvd}
\end{table}

\subsection{Analysis of learned boundaries}
\begin{figure}[tb]
\centering
\includegraphics[width=0.49\columnwidth]{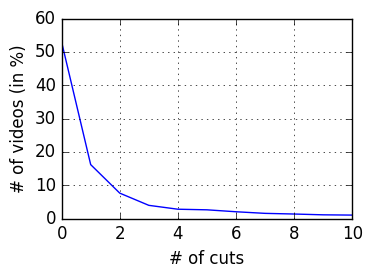}
\includegraphics[width=0.49\columnwidth]{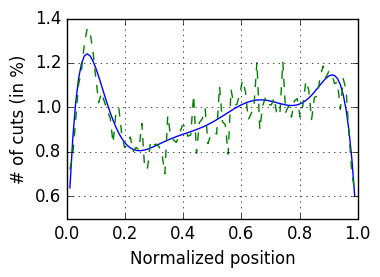}
\caption{Distribution of the number and position of detected cuts on the M-VAD and MPII-MD datasets. The dashed green line in the right plot shows the distribution of cuts with respect to their relative position inside the video (where 0 represents the beginning and 1 represents the end of a video) obtained with an histogram with 100 bins, while the solid blue line is obtained by fitting a polynomial with degree 10 on the histogram.}
\label{fig:charts}
\end{figure}

We collect statistics on the behavior of the boundary detector, which is the key component that rules the temporal structure of the video encoder. Figure~\ref{fig:charts} shows the distribution of the number and position of detected cuts on the M-VAD and MPII-MD datasets. As it can be observed, in the vast majority of the videos less than three boundaries are detected. This result is in contrast with the approach of~\cite{pan2015hierarchical}, in which the video was purposely segmented in very small chunks. Looking at the position of cuts, we also observe a linear growth in the probability of having a cut between the 20\% and 80\% of the duration of the video, so the more the video advances, the more the need of a cut increases. Two peaks can also be noticed, at the very beginning and ending of the video; this is due the fact that in the M-VAD and MPII-MD datasets videos are not precisely aligned with their captions, so the ends of the video are often uncorrelated with the main content of the video.

To confirm the effectiveness of the position of detected segments, we trained our network by forcing the encoder to split the input video in equally spaced chunks, maintaining the same number of segments detected by the original Boundary-aware encoder. This resulted in a reduction of $0.2\%$ METEOR on M-VAD, and $0.5\%$ METEOR on MPII-MD. 

We also compare the boundaries found by our neural model with those found by an off-the-shelf open source shot detector~\cite{baraldi2015shot}. Among all detected boundaries on the M-VAD and MPII-MD datasets, 33.7\% of them were found to be less than 15 frames far from a shot boundary. This confirms that the proposed LSTM cell can identify camera changes and appearance variations, but also detects more soft boundaries which do not correspond to shots.

Finally, we investigate how the the proposed video encoder would perform using shot boundaries detected with~\cite{baraldi2015shot} instead of those learned by the boundary detector. Results are reported in Tables~\ref{tab:mvad} and~\ref{tab:mpiimd}. On the M-VAD dataset, using shot boundaries resulted in a $7.1\%$ METEOR, which is $0.2\%$ below the performance of the Boundary-aware encoder, while on the MPII-MD dataset, we observed a $6.6\%$ METEOR, which again is below the result reported by our complete model. This confirms that, even though shots give a reasonable decomposition of the video, learned boundaries are definitely more effective and yield to better captioning performance.

\section{Conclusion}
In this work, we proposed a novel boundary-aware video encoder for the task of video captioning, which achieves competitive results across popular benchmarks. Our method can discover the hierarchical structure of the video, and modify the temporal connections of a recurrent layer accordingly. We believe that the proposed architecture is generic and could be employed in other video-related applications, such as video classification and action detection.

{\small
\section*{Acknowledgment}
This work has been partially founded by the project ``Citt\`a educante'' (CTN01\_00034\_393801) of the National Technological Cluster on Smart Communities (cofunded by the Italian Ministry of Education, University and Research - MIUR).
We acknowledge the CINECA award under the ISCRA initiative, for the availability of high performance computing resources and support. 
}

{\small
\bibliographystyle{ieee}
\bibliography{2017CVPR}
}

\end{document}